\documentclass[sigconf]{acmart}
\AtBeginDocument{%
  }

\usepackage{algorithm}
\usepackage{algorithmic}
\usepackage{multirow}

\copyrightyear{2025}
\acmYear{2025}
\setcopyright{acmlicensed}\acmConference[MM '25]{Proceedings of the 33rd ACM International Conference on Multimedia}{October 27--31, 2025}{Dublin, Ireland}
\acmBooktitle{Proceedings of the 33rd ACM International Conference on Multimedia (MM '25), October 27--31, 2025, Dublin, Ireland}
\acmDOI{10.1145/3746027.3755122}
\acmISBN{979-8-4007-2035-2/2025/10}

\newcommand{\modelname}{SVL\,}
\newcommand{\frameworkname}{Short-LVLM\,}

\newcommand{\noindentmodel}{SVL}
\newcommand{\noindentframework}{Short-LVLM}

\begin{document}
\begin{sloppypar}
\title{Short-LVLM: Compressing and Accelerating Large Vision-Language Models by Pruning Redundant Layers
}


\author{Ji Ma}
\authornote{Both authors contributed equally to this research.}
\affiliation{%
  %\institution{School of Computer Science and Ningbo Institute, Northwestern Polytechnical University}
    \institution{Northwestern Polytechnical University}
  \city{Xi’an, Shaanxi}
  \country{China}}
\affiliation{%
  \institution{National Engineering Laboratory for Integrated Aero-Space-Ground-Ocean Big Data Application Technology}
  \city{Xi’an, Shaanxi}
  \country{China}}
\email{maji@mail.nwpu.edu.cn}

\author{Wei Suo}
\authornotemark[1]
\authornote{Corresponding author}
\affiliation{%
  %\institution{School of Computer Science and Ningbo Institute, Northwestern Polytechnical University}
    \institution{Northwestern Polytechnical University}
  \city{Xi’an, Shaanxi}
  \country{China}}
\affiliation{%
  \institution{National Engineering Laboratory for Integrated Aero-Space-Ground-Ocean Big Data Application Technology}
  \city{Xi’an, Shaanxi}
  \country{China}}
\email{suowei1994@mail.nwpu.edu.cn}

\author{Peng Wang}
\affiliation{%
  %\institution{School of Computer Science and Ningbo Institute, Northwestern Polytechnical University}
    \institution{Northwestern Polytechnical University}
  \city{Xi’an, Shaanxi}
  \country{China}}
\affiliation{%
  \institution{National Engineering Laboratory for Integrated Aero-Space-Ground-Ocean Big Data Application Technology}
  \city{Xi’an, Shaanxi}
  \country{China}}
\email{peng.wang@nwpu.edu.cn}

\author{Yanning Zhang}
\affiliation{%
  %\institution{School of Computer Science and Ningbo Institute, Northwestern Polytechnical University}
    \institution{Northwestern Polytechnical University}
   \city{Xi’an, Shaanxi}
   \country{China}}
\affiliation{%
  \institution{National Engineering Laboratory for Integrated Aero-Space-Ground-Ocean Big Data Application Technology}
  \city{Xi’an, Shaanxi}
  \country{China}}
\email{ynzhang@nwpu.edu.cn}


\begin{abstract}
  Although large vision-language models (LVLMs) have demonstrated impressive capabilities in multi-modal understanding and reasoning, their practical applications are still limited by massive model parameters
  and high computational costs. 
  Recent efforts from natural language processing (NLP) have shown the effectiveness of layer pruning, offering a plausible training-free compression solution. 
  However, due to the modality divergence between vision and language, it is unclear whether these NLP techniques are still effective in LVLMs.
  In this paper, we empirically prove that directly applying these layer pruning methods to LVLMs is ineffective.
  Through extensive experiments, we find that non-essential vision-language (VL) tokens and inter-layer feature gaps pose critical challenges to pruning layers in LVLMs. Based on these insights, we propose a novel framework \frameworkname (\noindentmodel) that can utilize important VL tokens and mitigate the layer-wise feature gaps.
Notably, \frameworkname not only achieves a superior trade-off between performance and efficiency but also exhibits several potential advantages, \textit{i.e.,} training-free, model-agnostic, and highly compatible. The code for this work is publicly available at https://github.com/ASGO-MM/Short-LVLM.

\end{abstract}

\begin{CCSXML}
<ccs2012>
   <concept>
       <concept_id>10010147</concept_id>
       <concept_desc>Computing methodologies</concept_desc>
       <concept_significance>300</concept_significance>
       </concept>
   <concept>
       <concept_id>10010147.10010178.10010224</concept_id>
       <concept_desc>Computing methodologies~Computer vision</concept_desc>
       <concept_significance>500</concept_significance>
       </concept>
   <concept>
       <concept_id>10010147.10010178</concept_id>
       <concept_desc>Computing methodologies~Artificial intelligence</concept_desc>
       <concept_significance>300</concept_significance>
       </concept>
 </ccs2012>
\end{CCSXML}

\ccsdesc[300]{Computing methodologies}
\ccsdesc[500]{Computing methodologies~Computer vision}
\ccsdesc[300]{Computing methodologies~Artificial intelligence}

\keywords{Large Vision-Language Models, Model Compression, Layer Pruning}


\maketitle

\section{Introduction}
In recent years, large language models (LLMs) have achieved remarkable advancements in natural language processing (NLP)~\cite{llama,vicuna,qwen,qwen2}. This success has sparked growing interest in extending their capabilities to the multi-modal domains, leading to the emergence of large vision-language models (LVLMs)~\cite{llava,minigpt4,qwen-vl,mplug-owl2,suo2024rethinking,ma2024c3,suo2025octopus,suo2023s3c}. While LVLMs excel in handling diverse vision-language (VL) tasks, their practical applications remain constrained by huge model parameters and the associated computational overhead. 
Therefore, it is crucial to compress LVLMs and enhance their computational efficiency.

\begin{figure}[t]
  \centering
  \includegraphics[width=1\linewidth]{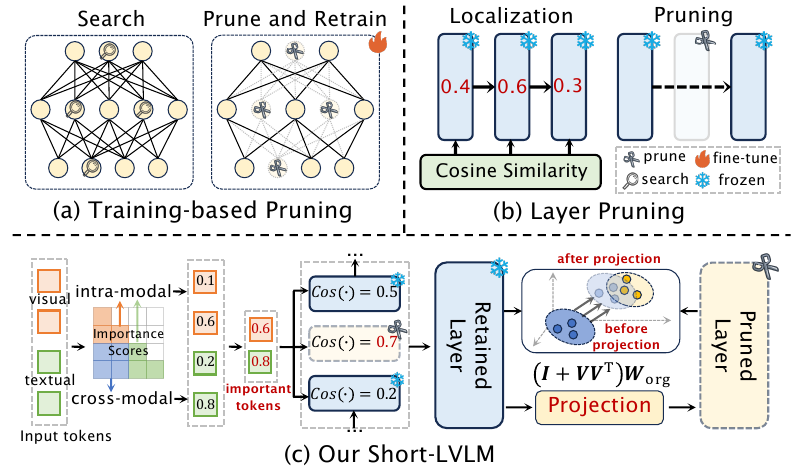}
  \caption{Comparison of different model pruning methods. (a) The training-based pruning methods involve fine-tuning the pruned models. (b) The layer pruning methods achieve training-free model compression via a \textit{localize-then-prune} process. (c) Our paradigm focuses on using important tokens for layer localization and 
 bridging inter-layer feature gaps via weight projection.
 }
  \label{fig:1}
\end{figure}

One prevalent approach for compressing LVLMs is via model pruning~\cite{upop,multi-flow,vl-lottery-tickets,mope-clip}, which aims to reduce the size of LVLMs by pruning redundant parameters. 
As shown in Fig.~\ref{fig:1}(a), the compressing workflow typically involves searching unnecessary weights and fine-tuning the pruned model to minimize performance degradation~\cite{upop,multi-flow,mope-clip,vl-lottery-tickets}. 
Despite the effectiveness, the fine-tuning process is both time-consuming and annotation-intensive, hindering the practicality of \textit{training-based pruning} methods in resource-constrained environments.
Fortunately, recent studies in NLP have revealed serious redundancy in the layers of  LLMs~\cite{sleb,shortgpt,gromov2024unreasonable}. 
As shown in Fig.\ref{fig:1}(b),  \textit{layer pruning} methods commonly employ a \textit{localize-then-prune} pipeline. By leveraging cosine similarity to identify redundant layers and directly remove them, LLMs can be effectively compressed in a training-free manner\cite{sleb,streamline-llm}.

Although these layer pruning techniques offer a plausible idea to compress LVLMs without retraining, their adaptation to the VL domain faces two significant challenges in localizing and pruning:
1) \textbf{Localizing:} 
 Relying solely on textual tokens to identify unnecessary layers as in NLP~\cite{sleb,streamline-llm} is biased and leads to suboptimal results (more discussions can be found in Sec.~\ref{sec:preliminary_layer_localizing}). Therefore, \textit{how to properly use VL tokens to localize redundant layers in LVLMs is an unsolved problem.}
2) \textbf{Pruning:}
LVLMs process vision and language modalities concurrently~\cite{llava-1.5,minigpt4}. Given the presence of the additional visual modality, \textit{it remains unclear whether directly pruning layers is still effective.}

In this paper, we first conduct exploratory experiments to  answer the aforementioned questions and offer the following new insights: 
1) We extend the layer pruning methods in NLP~\cite{shortgpt,gromov2024unreasonable} to the VL domain by utilizing all VL tokens to identify redundant layers. 
Interestingly, results indicate that the adaptation is ineffective.
We argue that this is due to the substantial redundancy in VL tokens~\cite{fastv,vtw,gamma-mod}, which introduces irrelevant information to disturb layer localization.
Consequently, \textit{identifying and utilizing important VL tokens is essential to localize redundant layers.}
2) 
By pruning layers from LVLMs and their language counterparts, 
we observe that LVLMs suffer more serious performance degradation than LLMs.
To further elucidate the underlying causes, we visualize the inter-layer feature similarity. 
The results show that LVLMs exhibit larger layer-wise feature gaps compared to LLMs, making them more sensitive to layer pruning.
Thus, \textit{it is crucial to devise a strategy for bridging the feature gaps.
}

Based on the above insights, we introduce a novel training-free framework called \textbf{S}hort-L\textbf{VL}M, abbreviated as \noindentmodel.
As illustrated in Fig.~\ref{fig:1}(c), \frameworkname leverages important tokens to localize redundant layers and reduces the feature gaps via weights projection.
Specifically, we first design \textbf{T}oken 
 \textbf{I}mportance \textbf{S}cores (TIS) to identify key tokens from both intra-modal and cross-modal perspectives.
 By computing cosine similarity using these important tokens, we can accurately prune the unnecessary layers in LVLMs.
Regarding layer-wise feature gaps, previous studies~\cite{residual_alignment,transformer_alignment} have demonstrated that the top singular vectors of different layers' Jacobian matrices are aligned in residual architectures, suggesting that feature gaps between layers can be represented using low-rank subspaces. Inspired by this, we propose \textbf{S}ubspace-\textbf{C}ompensated \textbf{P}runing (SCP), 
which leverages Singular Value Decomposition (SVD)~\cite{svd} to extract low-rank subspaces from the feature difference matrix between retained and pruned layers.
By projecting the weights of retained layers using the extracted subspaces, SCP effectively reconstructs the features of removed layers and restores the performance of LVLMs after pruning.

Benefiting from the above designs, \frameworkname not only effectively compresses LVLMs, but also offers several potential advantages: 
1) \textbf{Training-Free.} 
Unlike training-based pruning methods~\cite{upop,multi-flow,gamma-mod,roe-llava}, our \modelname performs pruning without fine-tuning LVLMs, offering an annotation-friendly and computationally efficient compression solution.
2) \textbf{Model-Agnostic.} 
Our method is designed to operate with minimal model-specific adjustments, allowing it to be effortlessly used on a wide range of LVLMs.
3) \textbf{Highly-Compatible.}  
Our \modelname can be easily combined with various model compression and acceleration techniques,~\textit{e.g.,}  model quantization~\cite{gptq} and token pruning~\cite{fastv}.

In summary, we make the following contributions:

\noindent $\bullet$ Our work makes the first attempt to prune LVLMs in a training-free manner.
We empirically prove that utilizing important VL tokens and bridging the inter-layer feature gaps are crucial for pruning layers in LVLMs.

\noindent $\bullet$ We propose a novel framework \frameworkname that can use important tokens for localization and restore the performance of pruned LVLMs
by bridging layer-wise feature gaps.
More importantly, our 
\modelname is training-free, model-agnostic and highly compatible.

\noindent $\bullet$ \frameworkname is evaluated on six prevalent multi-modal benchmarks with four LVLMs (from 7B to 13B).
The results show that \frameworkname can accelerate LLaVA-1.5-13B~\cite{llava-1.5} by 1.34$\times$ while preserving more than 96\% model capacity.

\section{Related Work}

\subsection{Model Pruning Methodologies}

Current model pruning methods for LLMs can be broadly divided into two categories: unstructured pruning and structured pruning.
Unstructured pruning~\cite{unstructed_pruning_1,unstructed_pruning_2,unstructed_pruning_3} works at the finer-grained weight level while structured pruning~\cite{structed_pruning_1,structed_pruning_2} is easier to achieve practical acceleration.
For vision-language models (VLMs), several works investigate pruning technologies. 
For example, Gan et al.~\cite{play_lottery_tickets} investigate the lottery ticket hypothesis in pre-trained VLMs, focusing on identifying sparse and trainable sub-networks via task-specific pruning.
Upop~\cite{upop} compresses VLMs by progressively searching and retraining pruned models in a joint optimization framework. 
More recently, 
MoPE-CLIP~\cite{mope-clip} performs pruning by measuring performance decline on cross-modal tasks.
MULTIFLOW~\cite{multi-flow} evaluates parameter importance based on magnitude and information flow to create a single pruned model transferable across multiple downstream tasks.
Our \frameworkname differs from these methods in two aspects: 1) As we set transformer layers as the basic pruning units, \frameworkname performs pruning without complex weight search algorithms. 2) More importantly, our method is training-free, offering an efficient approach for compressing LVLMs.

\subsection{Sparse Computations for LVLMs}

Instead of pruning model parameters to enhance computational efficiency, many efforts have been dedicated to sparse computations in LVLMs~\cite{fastv,vtw,gamma-mod,roe-llava,par}.
These approaches are grounded in the observation that many visual (or textual) tokens are irrelevant to a specific user instruction~\cite{fastv,gamma-mod}. By skipping computations for these unnecessary inputs, the inference process can be accelerated without sacrificing performance~\cite{vtw,roe-llava}. 
For example, ROE-LLaVA~\cite{roe-llava} utilizes a trained router and routing tokens to bypass computations in specific model layers.
Unlike these works, \frameworkname focuses on reducing the model parameters while simultaneously accelerating inference.
Furthermore, our approach is orthogonal to these techniques, meaning that \frameworkname can be seamlessly integrated with them to achieve an improved trade-off between performance and efficiency.

\section{Preliminary}
\label{sec:preliminary}
\subsection{Large Vision-Language Models}

Given an image $I$ and a user's textual query $Q$, LVLMs are capable of handling a wide range of vision-language (VL) tasks. After encoding \( I \) and \( Q \) into token sequences \( [v_1, \dots, v_s] \) and \( [q_1, \dots, q_c] \), the models can use them to perform auto-regressive generation. This process is formulated as:

\begin{equation}
    p(y) = \prod \limits_{i=1}^np_\theta(y_i|{v_1},...\;,{v_s},q_1,...\;,q_c,y_{<i}).
    \label{eq:1}
\end{equation}

where $s,c$ and $n$ denote the length of visual tokens, textual tokens and model outputs. 
By leveraging extensive training to extend the capabilities of LLMs to the vision-language domain, LVLMs excel in multi-modal understanding and reasoning~\cite{llava,llava-1.5,qwen-vl}. However, their practical applications are constrained by the massive model parameters and the associated computational overhead.

To tackle this problem, previous training-based pruning efforts~\cite{upop,multi-flow,gamma-mod,roe-llava} apply a \textit{prune-then-retrain} approach,  mitigating the performance degradation via fine-tuning the pruned LVLMs.
Despite their effectiveness, this approach demands extensive annotations and incurs significant training costs (\textit{e.g.,} RoE-LLaVA~\cite{roe-llava} requires $\sim$100 GPU hours to fine-tune the pruned model). Recent studies~\cite{shortgpt,sleb,gromov2024unreasonable} reveal that LLMs suffer from severe layer redundancy. By removing the less important layers, LLMs can be effectively compressed without retraining. Since LVLMs are built upon the LLMs~\cite{interprete_vl_process},  
a natural idea to achieve training-free compression for LVLMs would be adapting these technologies to the VL domain.
Next, we will introduce these NLP methodologies~\cite{shortgpt,sleb,gromov2024unreasonable} and discuss the challenges of adapting them to LVLMs.

\begin{figure}[t]
  \centering
  \includegraphics[width=1\linewidth]{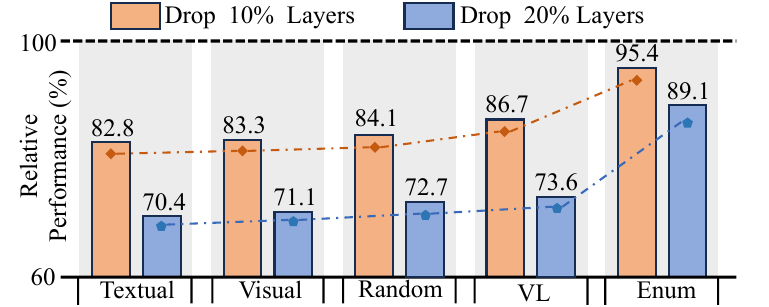}
  \caption{
  Preliminary experiments using different tokens (\textit{e.g.,} visual, textual, and VL) to localize redundant layers in LLaVA-1.5-7B.
  The “Relative Performance” is the relative performance compared to the original LVLM, \textit{i.e.,} the full model without pruning.
“Random” denotes randomly dropping layers, and “Enum” is enumeration-based search (can be regarded as the performance upper bound).
The results indicate that using either uni-modal or VL tokens is ineffective for localizing redundant layers in LVLMs.}
  \label{fig:layer_localization}
  
\end{figure}

\subsection{Layer Pruning Methodologies for LLMs}
\label{sec:layer_pruning_4_llm}
Current LLMs~\cite{llama,qwen,vicuna} generally utilize a stack of Transformer layers with residual connections. 
Assuming the LLM consists of $L$ layers, the output logits can be expressed as ~\cite{gromov2024unreasonable,sleb}:
\begin{equation}
   \boldsymbol{x}^{(L)}=\boldsymbol{x}^{(0)}+\sum_{i=0}^{L-1}f(\boldsymbol{x}^{(\ell)},\theta^{(\ell)}),
    \label{eq:total_layers_evolution}
\end{equation}
where $\boldsymbol{x}^{(\ell)}$ and $\theta^{(\ell)}$ are hidden states and parameters of layer $\ell$. $f$ is the transformation applied by layer $\ell$. 
From Eq.~\ref{eq:total_layers_evolution}, the outputs of LLMs can be decomposed into contributions from each layer. Given the conditions that these contributions are \textit{independent} and $L\gg 1$, the removal of certain layers would have a negligible impact~\cite{gromov2024unreasonable}.
Based on this conclusion, previous NLP methodologies~\cite{sleb,streamline-llm} localize the redundant layers using cosine similarity:
\begin{equation}
cos(\boldsymbol{x}^{(\ell)},\boldsymbol{x}^{(\ell+1)})=\mathbb E_{(\boldsymbol{x}^{(\ell)},\boldsymbol{x}^{(\ell+1)})\in\mathcal D}(\frac{\boldsymbol{x}^{(\ell)}\cdot \boldsymbol{x}^{(\ell+1)}}{||\boldsymbol{x}^{(\ell)}||\cdot ||\boldsymbol{x}^{(\ell+1)}||}),
\label{eq:cos_sim}
\end{equation}
Here, $\mathcal{D}$ denotes the calibration data. By calculating $cos(\cdot)$ for all layers $\ell\in [1,L]$ and removing those with high similarity, LLMs can be effectively pruned without further fine-tuning.

However, considering that LVLMs process visual and linguistic modalities concurrently~\cite{llava-1.5,minigpt4}, several questions remain to be answered before transferring these techniques to the VL domain: 1) \textbf{Localizing:} 
Given the presence of the additional visual modality, identifying the redundant layers is more challenging compared to the uni-modal scenarios.
\textit{How to properly use VL tokens to localize redundant layers is an open question.} 2) \textbf{Pruning:} 
In LVLMs, features $\boldsymbol{x}$ in each layer are enriched by the visual modality. Due to the inherent gaps between vision and language~\cite{hacl}, \textit{it remains unclear whether directly pruning layers in LVLMs would still be effective.}

To answer the above questions, we conduct related experiments on two popular VL models, \textit{i.e.,} LLaVA-1.5-7B~\cite{llava-1.5} and Qwen-VL-Chat-7B~\cite{qwen}. 
For LVLMs, to ensure the conclusions are generalizable, four prevalent multi-modal benchmarks are used, \textit{i.e.,} ScienceQA~\cite{scienceqa}, SEED-Bench~\cite{seed-bench}, POPE~\cite{pope} and MM-Bench~\cite{mmbench}. For their language counterparts, we follow previous LLMs pruning works~\cite{sleb,shortgpt,streamline-llm} and use natural language understanding benchmarks.

\begin{figure}[t]
  \centering
  \includegraphics[width=1\linewidth]{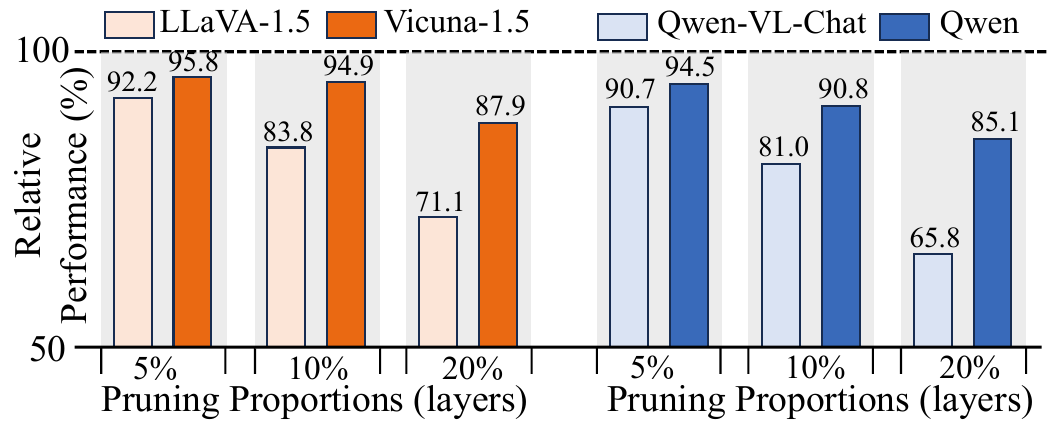}
  
  \caption{Preliminary experiments exploring the effectiveness of directly removing layers in LVLMs.
    The “Relative Performance” is the relative performance compared to the full model without pruning.
  The results show that LVLMs are more sensitive to layer pruning compared to their language counterparts. 
  }
\label{fig:layer_pruning}
\end{figure}

\begin{figure}[t]
  \centering
  \includegraphics[width=1\linewidth]{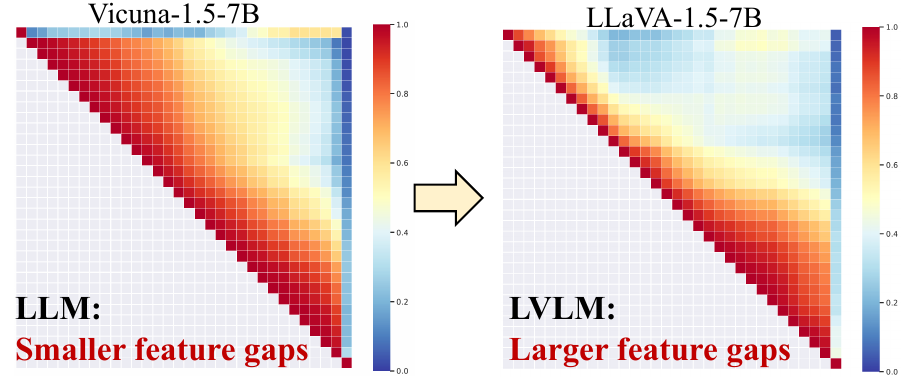}
  \caption{Visualization of feature similarity across layers. 
  Warm color (close to red) denotes high similarity.
  LLaVA-1.5-7B and its language counterparts (\textit{i.e.,} Vicuna-1.5-7B) are used to conduct the experiments.
  The results show that LVLMs exhibit larger inter-layer feature gaps compared to LLMs. 
  }
\label{fig:layer_pruning_vis}
\end{figure}

\subsection{Layer Localizing Analyses for LVLMs}
\label{sec:preliminary_layer_localizing}
In this section, we aim to explore how VL tokens can be leveraged to effectively localize the less important layers.
Specifically, we extend the NLP technologies in Sec.~\ref{sec:layer_pruning_4_llm}~to the VL domain by computing cosine similarity with textual, visual, and VL tokens, respectively. 
By observing the performance degradation caused by pruning layers identified as “redundant” by these different tokens, we can determine which type of tokens more accurately measures the layer redundancy of LVLMs.
Moreover, to establish a reference, we further adopt an enumeration-based search to explore all possible layer-pruning combinations under predefined pruning proportions (\textit{e.g.,} 10\% or 20\%).
By testing models with enumerated pruning configurations, we are able to find the upper bound of performance when a specific proportion of layers are removed.

As shown in Fig.~\ref{fig:layer_localization}, the relative performance of the pruned models is reported (the original model without pruning is 100\%). 
From the figure, relying on uni-modal tokens (\textit{i.e.,}  textual or visual) to localize unnecessary layers leads to significant performance degradation, with the removal of 20\% layers causing approximately a 30\% drop in performance.
Notably, these results are worse than randomly dropping layers (\textit{i.e.,} plots with x-axis “Random”), further demonstrating their ineffectiveness in localizing redundant layers.
Since LVLMs process vision and language modalities simultaneously, \textit{assessing layer redundancy in a single modality (textual or visual) inevitably neglects information from the other, thus resulting in suboptimal outcomes.}

To address this limitation, a natural idea is to combine all vision-language (VL) tokens and identify the less important layers based on the information from both modalities. 
However, as illustrated in the plot with the x-axis labeled “VL,” this approach yields only marginal improvements compared to the uni-modal scenarios. Furthermore, we observe a significant performance discrepancy between “VL” and “Enum” (\textit{i.e.,} enumeration-based searches). 
These results indicate that \textit{simply utilizing all VL tokens to identify redundant layers in LVLMs is useless}, despite its effectiveness in NLP~\cite{sleb,streamline-llm}.

A plausible explanation for this phenomenon lies in the inherent redundancy of VL tokens~\cite{gamma-mod,roe-llava}, especially the visual tokens~\cite{fastv,vtw}.
In fact, when directly applying all tokens to compute cosine similarity and localize redundant layers, a substantial number of irrelevant tokens inevitably introduce noisy information, thereby hindering the accuracy of layer localization.

Based on the above analysis, we conclude that: Unlike layer pruning methods in NLP that typically employ all tokens to localize redundant layers, \textbf{\textit{identifying and utilizing important tokens is essential for localizing redundant layers in LVLMs.}}

\begin{figure*}[t]
  \centering
  \includegraphics[width=1\linewidth]{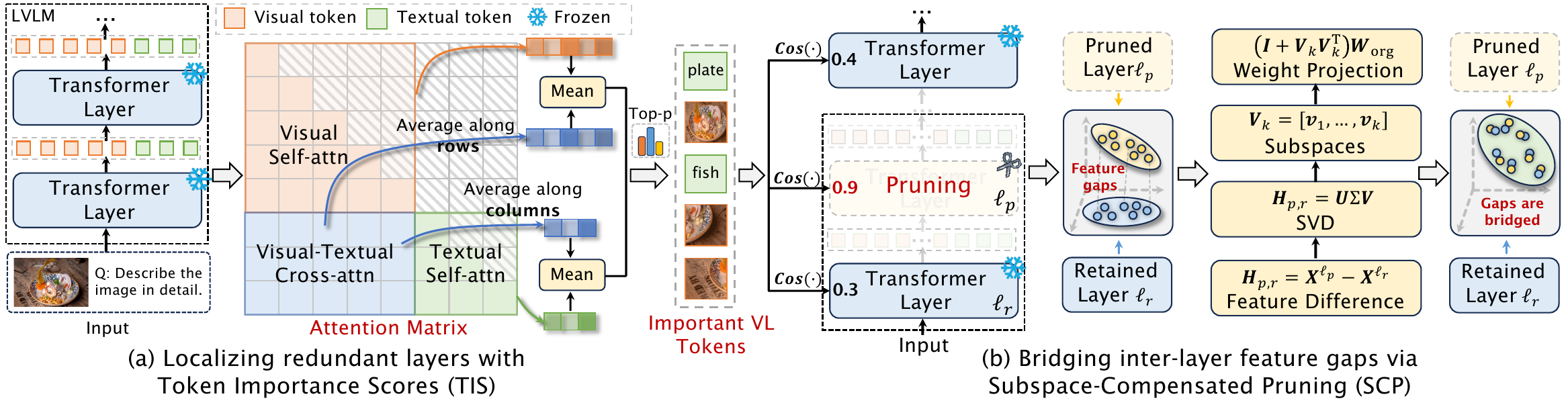}
  \caption{Overview of our \frameworkname (\noindentmodel). Our \frameworkname consists of two key designs: (a) Localizing redundant layers with Token Importance Scores (TIS) and (b) Bridging inter-layer feature gaps via Subspace-Compensated Pruning (SCP).}
  \label{fig:model}
\end{figure*}

\subsection{Layer Pruning Analyses for LVLMs}

\label{sec:preliminary_layer_removing}

The analyses in Sec.~\ref{sec:preliminary_layer_localizing} demonstrate that key VL tokens play a vital role in localizing redundant layers. In this section, we focus on the other critical problem, \textit{i.e., layer pruning}: whether directly pruning layers without further post-processing is still effective in LVLMs.

To counteract this, we randomly prune the same layers from LVLMs 
and their language counterparts multiple times. 
By comparing the performance degradation between LVLMs and LLMs, we try to answer whether directly pruning layers is still effective in LVLMs. 
Specifically, as illustrated in Fig.~\ref{fig:layer_pruning}, the experiments are conducted with pruning proportions of 5\%, 10\%, and 20\%. 
For each proportion, we repeat pruning with
ten different seeds, leading to ten unique pruning results. By evaluating the pruned models and averaging the results across ten seeds, we derive the overall impact of layer pruning on model performance.
From the figure, it can be observed that the performance of both LVLMs and LLMs exhibits evident declines as the proportion of pruned layers increases.
However, the performance degradation of LVLMs is more serious compared to
 their language counterparts, particularly at high pruning proportions. For instance, when the pruning proportion is 20\%, the relative performance of Qwen-VL-Chat is 65.8\% while Qwen is 85.1\%, showing a gap of 20 percentage points. Meanwhile, we observe that a similar pattern also exists in LLaVA-1.5 and its language counterpart Vicuna-1.5.
Overall, the above results indicate that \textit{directly pruning layers without post-processing is NOT suitable for LVLMs.} 

To investigate the underlying causes, we visualize the inter-layer feature similarity of~LLaVA-1.5-7B~\cite{llava-1.5} and its language counterpart Vicuna-1.5-7B~\cite{vicuna}.
As shown in Fig.~\ref{fig:layer_pruning_vis}, warm colors (\textit{i.e.,} closer to red) indicate higher similarity, and vice versa.
From the figure, we can find that LVLMs exhibit larger layer-wise feature gaps compared to their language counterparts, making them more sensitive to layer pruning.

Based on the above experiments, we conclude that \textbf{\textit{directly removing layers is NOT effective due to the layer-wise feature gaps in LVLMs, thus it is important to develop a strategy for bridging these feature gaps.}}

\section{Method}

From the analyses in Sec.~\ref{sec:preliminary}, it can be found that 1) \textbf{Layer Localizing:} Naively combining all VL tokens offers limited value. Conversely, leveraging important VL tokens to reduce interference from irrelevant information aids in identifying redundant layers. 2) \textbf{Layer Pruning:} Due to the inter-layer feature gaps in LVLMs, directly pruning layers proves ineffective. Therefore, developing a strategy to bridge the feature gaps is essential for alleviating the performance degradation of the pruned model.

Based on these insights, we propose a simple yet effective framework \textbf{S}hort-L\textbf{VL}M (SVL), which can localize redundant layers with \textbf{T}oken \textbf{I}mportance \textbf{S}cores (TIS) and bridge the layer-wise feature gaps via \textbf{S}ubspace-\textbf{C}ompensated \textbf{P}runing (SCP). 
As shown in Fig.~\ref{fig:model}, rather than relying on complex weight search algorithms~\cite{upop,multi-flow} or extensive fine-tuning of pruned LVLMs~\cite{roe-llava,gamma-mod}, our method uses layers as the basic dropping units and performs weight projection to restore performance.
Next, we will introduce \frameworkname in detail.

\subsection{Token Importance Scores}

\label{sec:method_layer_localization}

As analyzed in Sec.~\ref{sec:preliminary_layer_localizing}, identifying and using important VL tokens are crucial for accurately localizing redundant layers. As shown in Fig.~\ref{fig:model}(a), we design Token Importance Scores (TIS) that identify key tokens from both intra-modal and cross-modal perspectives. 

Specifically, given the calibration data \(\mathcal{D}\), we use $d_i = \langle V, Q \rangle$ as an example to illustrate our method. By feeding the visual input \(V = [v_1, \dots, v_s]\) and textual query \(Q = [q_1, \dots, q_c]\) into the LVLMs, the models can output
the responses auto-regressively. At layer \(\ell\), the corresponding hidden states of these VL tokens are obtained:
\begin{equation}
\boldsymbol{V}^{\ell}=[\boldsymbol{v}_1^{\ell},...\;,\boldsymbol{v}_s^{\ell}],~
\boldsymbol{Q}^{\ell}=[\boldsymbol{q}_1^{\ell},...\;,\boldsymbol{q}_c^{\ell}],
\end{equation}
where $s$ and $c$ denote the length of visual and textual tokens.  Since both vision and language modalities are crucial for vision-language (VL) tasks~\cite{llava-1.5,qwen-vl}, for each token, its importance is measured from two perspectives: intra-modal and cross-modal. 
Previous studies~\cite{fastv,ATP-LLaVA} have demonstrated that attention scores can serve as strong indicators for token importance. 
Building on this insight, taking 
$\boldsymbol{v}_i^{\ell}$ as an example, we first compute the self-attention scores it receives and use them as intra-modal importance:
\begin{equation}
S_i^{\rm intra} = \frac{1}{s}\sum_{j=1}^{s}{\rm SelfAttn}(\boldsymbol{v}_i^{\ell},\boldsymbol{v}_j^{\ell}),
\end{equation}
where SelfAttn denotes the self-attention operation in the models.
Then, the cross-modal importance for $\boldsymbol{v}_i^{\ell}$ is computed as below:
\begin{equation}
S_i^{\rm cross} = \frac{1}{c}\sum_{j=1}^{c}{\rm CrossAttn}(\boldsymbol{v}_i^{\ell},\boldsymbol{q}_j^{\ell}).
\end{equation}
Here, CrossAttn denotes the cross-attention operations between visual and textual modalities.
Finally, by averaging these two scores and repeating the computations for all $\boldsymbol{v}_i^{\ell}$, where $i\in [1,...\;,s]$, we can obtain the TIS for all visual tokens at layer $\ell$. 
Similarly, the TIS for textual tokens can be derived by applying the same computation process for all $\boldsymbol{q}_i^{\ell}$, where $i\in [1,...\;,c]$. 

After calculating the TIS for all VL tokens at layer $\ell$, we sort them based on their values and perform a Top-$p$ sampling~\cite{top-p} to discard the non-essential tokens.
Here, $p$ is the sampling ratio (\textit{e.g.,} 10\%). 
With irrelevant information removed, the redundancy of layer $\ell$ can be effectively evaluated by computing the cosine similarity on the key tokens. By repeating the above procedure across all model layers, we can accurately localize less important layers and prune them based on the predefined sparsity threshold.

\subsection{Subspace-Compensated Pruning}

Given a predefined layer sparsity, we can identify and remove less important layers as described above, resulting in the pruned layer set $L_p$ and the retained layer set $L_r$. However, due to the layer-wise feature gaps in LVLMs, even when these less important layers are accurately identified and removed, the performance drop of pruned models remains substantial (more discussions can be found in Sec.~\ref{sec:preliminary_layer_removing}). Therefore, as shown in Fig.~\ref{fig:model}(b), we propose Subspace-Compensated Pruning (SCP) to alleviate this problem.

Previous studies~\cite{residual_alignment,transformer_alignment} have shown that the top singular vectors of Jacobian matrices across different layers are aligned in residual architectures, with this alignment becoming more evident when the layers are adjacent or closely positioned.
This observation indicates that the feature space of one layer is highly representable by another (especially when their locations are near), thus feature gaps between them can be effectively captured using low-rank subspaces. 
Inspired by this, we employ Singular Value Decomposition (SVD)~\cite{svd,nullu} to extract low-rank subspaces from the difference matrix between the features of retained and pruned layers. By projecting the weights of the retained layers using the extracted subspaces, SCP effectively reconstructs the features of pruned layers and restores the performance of LVLMs after pruning.

In particular, for the pruned layer set $\displaystyle L_p$, we use $\displaystyle \ell_{p}$ in $\displaystyle L_p$ as an example to introduce our method. 
As described above, layers that are closer in position exhibit better feature alignment. Therefore, we choose $\displaystyle \ell_{r}$ in $\displaystyle L_r$ as the candidate layer, where $\ell_{r} = \arg\min |\ell_{r} - \ell_{p}|$.
Since the model performance collapses when the pruning ratio exceeds 50\% in our experiments, we maintain the pruning ratio below this threshold. Benefiting from this configuration, for each $\displaystyle \ell_{p}$, there is an available layer in $\displaystyle L_r$ to act as the candidate layer. 
In practice, multiple layers in $\displaystyle L_p$ may correspond to the same $\displaystyle \ell_{r}$. 
To prevent potential feature conflicts, we remove the selected $\displaystyle \ell_{r}$ from $\displaystyle L_r$, ensuring each layer in $\displaystyle L_r$ can only be chosen once.
After finding the candidate layer $\displaystyle \ell_{r}$ for $\displaystyle \ell_{p}$, we utilize the calibration data $\displaystyle \mathcal{D}$ to evaluate the layer-wise feature gaps. By extracting and stacking representations of the data samples, their hidden states $\displaystyle \boldsymbol{X}^{\ell}\in \mathbb R^{N\times D}$ at layer $\displaystyle \ell$ can be obtained, where $N$ is the number of data samples and $D$ is the hidden dimensions.
Then, the feature difference matrix of $\displaystyle \ell_{p}$ and $\displaystyle \ell_{r}$ can be calculated as:
\begin{equation}
    \boldsymbol{H}_{p,r}=\boldsymbol{X}^{\ell_{p}}-\boldsymbol{X}^{\ell_{r}}.
\end{equation}
For  $\boldsymbol{H}_{p,r}\in \mathbb R^{N\times D}$, we leverage Singular Value Decomposition (SVD)~\cite{svd} to extract low-rank subspaces.
The calculation process is formulated as:
\begin{equation}
    \boldsymbol{H}_{p,r}=\boldsymbol{U}\boldsymbol{\Sigma}\boldsymbol{V},
\end{equation}
where  $\boldsymbol{U}\in\mathbb R^{N\times N}$,$\boldsymbol{V}\in \mathbb R^{D\times D}$ and $\boldsymbol{\Sigma}\in \mathbb R^{N\times D}$ is a diagonal matrix with singular values arranged in descending order.
Next, we select \(k\) columns from \(\boldsymbol{V}\), \textit{i.e.,} the singular vectors associated with the top-\(k\) singular values to form the subspaces: 
\begin{equation}
    \boldsymbol{V}_k=[\boldsymbol{v}_1,...\;,\boldsymbol{v}_k],
\end{equation}
where $\boldsymbol{V}_k\in \mathbb R^{D\times k}$ and $k\ll D$.
Finally, we project the weights of the retained layer $\ell_{r}$ using $\boldsymbol{V}_k$:
\begin{equation}
    \boldsymbol{W}^{\ell_{r}}_{\rm proj}=(\boldsymbol{I}+\boldsymbol{V}_k\boldsymbol{V}_k^\top)\;\boldsymbol{W}^{\ell_{r}}_{\rm org}.
\end{equation}
The corresponding forward process of layer $\ell_{r}$ after weights projection is as follows:
\begin{equation}
\begin{aligned}
\label{eq:forward_after_proj}
\boldsymbol{x}^{\ell_{r}}_{\rm proj}&= \boldsymbol{x}^{\ell_{r}}_{\rm org}(\boldsymbol{I}+\boldsymbol{V}_k\boldsymbol{V}_k^\top)\;\boldsymbol{W}^{\ell_{r}}_{\rm org}\\
&=\underbrace{\boldsymbol{x}^{\ell_{r}}_{\rm org}\boldsymbol{W}^{\ell_{r}}_{\rm org}}_{\text{original feature}} + \underbrace{\boldsymbol{x}^{\ell_{r}}_{\rm org}\boldsymbol{V}_k\boldsymbol{V}_k^\top\boldsymbol{W}^{\ell_{r}}_{\rm org}}_{\text{subspace feature}}.
\end{aligned}
\end{equation}
Since SVD provides the optimal low-rank approximation of the feature difference matrix \(\boldsymbol{H}_{p,r}\), the features of pruned layer $\ell_{p}$ can be effectively reconstructed using the \textit{subspace feature} term, thereby bridging the features gaps and restoring the performance of the pruned LVLMs.

\section{Experiment}

\subsection{Experimental Settings}

\subsubsection{Benchmarks and Backbones.}
We conduct extensive experiments on six multi-modal benchmarks, \textit{i.e.,} AOKVQA~\cite{aokvqa}, ScienceQA-Img~\cite{scienceqa}, MME~\cite{mme}, POPE~\cite{pope}, MMBench~\cite{mmbench} and SEED-Bench-Img~\cite{seed-bench}. 
We use the default metrics and settings of these benchmarks to conduct all experiments.
Moreover, we apply \frameworkname on four prevalent LVLMs, \textit{i.e.,} LLaVA-1.5-7B~\cite{llava-1.5}, mPLUG-Owl2-7B~\cite{mplug-owl2}, Qwen-VL-Chat-7B~\cite{qwen-vl} and LLaVA-1.5-13B~\cite{llava-1.5}.

\subsubsection{Implementation Details.} 
To balance performance and efficiency, we use the calibration data $\mathcal{D}$ with randomly sampled 1k instances from LLaVA-665k~\cite{llava-1.5}. 
The $p$ of Top-$p$ sampling~\cite{top-p} for discarding non-essential tokens is set as 10\% and the rank $k$ of the low-dimensional subspace extracted using SVD is set as 64 by default.
As previous works~\cite{gromov2024unreasonable,vtw} have shown that the latter half of models tends to exhibit more serious redundancy, we focus the pruning process on these deeper layers (\textit{e.g.,} the last 16 layers of LLaVA-1.5-7B~\cite{llava-1.5}).

\begin{table*}[t]
\centering
\caption{
Results of \frameworkname under varying \textit{Pruning Ratios} across different \textit{Model Architectures} and \textit{Model Scales}. \textit{Acc} and \textit{Speed} denote the \textit{accuracy} and \textit{samples per second}, respectively.
}
\label{tab:main_results}

\begin{tabular}{l|cc|cc|cc|cc|cc|cc|cc}
\toprule
\multicolumn{1}{c|}{\multirow{2}{*}{\textbf{Methods}}} & \multicolumn{2}{c|}{\textbf{AOKVQA}} & \multicolumn{2}{c|}{\textbf{ScienceQA$\rm^I$}} & \multicolumn{2}{c|}{\textbf{MME}} & \multicolumn{2}{c|}{\textbf{POPE}}  & 
\multicolumn{2}{c|}{\textbf{MMBench}}  &
\multicolumn{2}{c|}{\textbf{SEED-Bench$\rm^I$}}  & \multicolumn{2}{c}{\textbf{Average}} \\
\multicolumn{1}{c|}{} & \multicolumn{1}{c}{Acc}  & \multicolumn{1}{c|}{Speed} & \multicolumn{1}{c}{Acc} & \multicolumn{1}{c|}{Speed} & \multicolumn{1}{c}{Acc} & \multicolumn{1}{c|}{Speed} & \multicolumn{1}{c}{Acc}& \multicolumn{1}{c|}{Speed}  & 
\multicolumn{1}{c}{Acc}& \multicolumn{1}{c|}{Speed}  & 
\multicolumn{1}{c}{Acc}& \multicolumn{1}{c|}{Speed}  & 
\multicolumn{1}{c}{Acc} &
\multicolumn{1}{c}{Speed}\\ 
\midrule
 LLaVA-1.5-7B~\cite{llava-1.5}  
 & 77.9 & 4.92 & 65.6 & 4.85 & 1246.4 & 5.27 & 84.8 & 3.45 & 64.3 & 4.81 & 65.4 & 5.19 & 70.1 & 4.75 \\ 
+\noindentframework$_{10\%}$ & 78.0 & 5.10 & 65.5 & 5.07 & 1219.6 & 5.45 & 84.8 & 3.62 & 65.1 & 5.09 & 65.2 & 5.35 & 69.9 & 4.95  \\
+\noindentframework$_{20\%}$ & 77.8 & 5.63 & 64.7 & 5.60 & 1202.9 & 6.03 & 84.6 & 3.97 & 64.3 & 5.66 & 64.6 & 5.84 & 69.4 & 5.46 \\
+\noindentframework$_{30\%}$ & 77.6 & 6.11 & 63.6 & 5.87 & 1147.0 & 6.47 & 84.1 & 4.26 & 64.2 & 6.10 & 63.1 & 6.49 & 68.4 & 5.88  \\
  \midrule
Qwen-VL-Chat-7B~\cite{qwen-vl}  
& 74.2 & 4.11 & 67.4 & 3.82 & 1509.5 & 3.90 & 86.4 & 2.95 & 65.3 & 3.47 & 65.5 & 3.73 & 72.4 & 3.66 \\
+\noindentframework$_{10\%}$ & 73.9 & 4.32 & 67.0 & 4.08 & 1484.6 & 4.11 & 86.1 & 3.19 & 64.3 & 3.67 & 65.3 & 3.93 & 71.8 & 3.88 \\
+\noindentframework$_{20\%}$ & 73.6 & 4.71 & 66.3 & 4.37 & 1467.5 & 4.42 & 85.2 & 3.36 & 63.8 & 4.05 & 64.4 & 4.35 & 71.1 & 4.21  \\
+\noindentframework$_{30\%}$ & 73.1 & 5.13 & 65.1 & 4.82 & 1302.0 & 4.97 & 83.1 & 3.64 & 62.6 & 4.20 & 63.7 & 4.71 & 68.8 & 4.58 \\
 \midrule

mPLUG-Owl2-7B~\cite{mplug-owl2}  
& 79.7 & 4.58 & 67.3 & 4.15 & 1394.3 & 4.32 & 85.3 & 3.07 & 64.6 & 4.12 & 65.0 & 4.80 & 71.9 & 4.17 \\
+\noindentframework$_{10\%}$ & 79.6 & 4.81 & 66.7 & 4.29 & 1393.8 & 4.53 & 85.2 & 3.22 & 64.5 & 4.26 & 64.8 & 5.00 & 71.8 & 4.35 \\
+\noindentframework$_{20\%}$ & 79.6 & 5.31 & 66.5 & 4.79 & 1388.5 & 4.92 & 85.1 & 3.45 & 64.3 & 4.55 & 64.5 & 5.48 & 71.6 & 4.75 \\
+\noindentframework$_{30\%}$ & 79.5 & 5.61 & 66.1 & 5.07 & 1384.2 & 5.34 & 84.8 & 3.82 & 64.2 & 5.19 & 64.4 & 5.93 & 71.4 & 5.16 \\
\midrule
LLaVA-1.5-13B~\cite{llava-1.5} & 82.5 & 3.47 & 70.1 & 3.14 & 1531.3 & 4.42 & 85.9 & 2.75 & 67.7 & 4.10 & 67.6 & 3.98 & 75.1 & 3.64 \\ 
+\noindentframework$_{10\%}$ & 82.3 & 3.61 & 69.9 & 3.23 & 1510.7 & 4.76 & 85.9 & 2.86 & 67.4 & 4.33 & 67.6 & 4.21 & 74.8 & 3.83 \\
+\noindentframework$_{20\%}$ & 82.0 & 3.85 & 69.7 & 3.60 & 1464.2 & 5.04 & 85.8 & 3.16 & 66.8 & 4.85 & 67.4 & 4.30 & 74.2 & 4.13 \\
+\noindentframework$_{30\%}$ & 81.4 & 4.17 & 69.4 & 3.92 & 1431.8 & 5.62 & 85.5 & 3.47 & 66.5 & 5.25 & 67.1 & 4.83 & 73.6 & 4.54 \\
+\noindentframework$_{40\%}$ & 80.6 & 4.41 & 69.3 & 4.18 & 1336.4 & 6.07 & 85.1  & 3.53 & 65.8 & 5.59 & 66.9 & 5.36 & 72.4 & 4.86 \\
\bottomrule
\end{tabular}
\end{table*}

\begin{table}[t]
\centering
\caption{
Ablation study. We ablate key components to demonstrate the effectiveness of our method. “Ratios” and “Speed” denote the \textit{pruning ratios} and \textit{samples per second}, respectively. “TIS” denotes the Token Importance Scores, while “SCP” denotes the Subspace-Compensated Pruning.
}
\label{tab:ablation_study}

\setlength\tabcolsep{3pt}
\begin{tabular}{c|cc|cc|c}
\toprule
\textbf{Ratios}  & \textbf{TIS} & \textbf{SCP} & \textbf{AOKVQA} & \textbf{MMBench} & \textbf{Speed}  \\

\midrule
0\% (\textit{full model}) & & & 77.9 & 64.3 & 4.87 \\
\midrule
%\midrule
10\% (\textit{baseline}) & 
& & 76.2 & 62.9 & 5.10  \\
10\% & \checkmark & & 77.4 & 63.6 & 5.10 \\
10\% & & \checkmark & 77.6 & 64.1 & 5.10 \\
10\% & \checkmark & \checkmark & 78.0 & 65.1 & 5.10  \\
\midrule
20\% (\textit{baseline}) & & & 75.5 & 61.4 & 5.65 \\
20\% & \checkmark & & 76.1 & 62.5 & 5.65 \\
20\% & & \checkmark & 77.0 & 63.2 & 5.65 \\
20\% & \checkmark & \checkmark & 77.8 & 64.3 & 5.65 \\
\midrule
30\% (\textit{baseline}) & & & 74.9 & 60.1 & 6.11 \\
30\% & \checkmark & & 76.4 & 62.9 & 6.11 \\
30\% & & \checkmark & 75.8 & 62.0 & 6.11 \\
30\% & \checkmark & \checkmark & 77.6 & 64.2 & 6.11 \\
\bottomrule
\end{tabular}
\end{table} 

\subsection{Quantitative Evaluation}

\subsubsection{Main Results}

In Table~\ref{tab:main_results}, we evaluate the effectiveness of \frameworkname across three \textit{model architectures} (\textit{i.e.,} LLaVA-1.5-7B~\cite{llava-1.5}, Qwen-VL-Chat-7B~\cite{qwen-vl} and mPLUG-Owl2-7B~\cite{mplug-owl2}) and two \textit{model scales} (\textit{i.e.,} LLaVA-1.5-7B~\cite{llava-1.5} and LLaVA-1.5-13B~\cite{llava-1.5}). From the table, it can be observed that our \modelname provides an effective trade-off between performance and efficiency across six benchmarks. 
In particular, \noindentframework$_{30\%}$ accelerates inference speed by $1.23\times$, with less than a 2\% performance decline. Furthermore, the effectiveness of our \modelname generalizes well across different model architectures.
Considering the mPLUG-Owl2-7B, \noindentframework$_{30\%}$ significantly compresses the model and enhances computational efficiency with negligible performance degradation (\textit{i.e.,} only 0.5\% averaged performance drops across six benchmarks).
As demonstrated by scaling lows~\cite{scaling_laws}, LVLMs with more parameters typically exhibit better multi-modal capacity. Therefore, we also evaluate \frameworkname on LLaVA-1.5-13B.
Notably, we find that this larger model can undergo more aggressive pruning (40\% pruning ratio) with little performance drop (only 2.7\%). 
The above results demonstrate that our \modelname is agnostic to both model architectures and scales. Additionally, different pruning ratios can be set for \frameworkname to tackle various practical scenarios.

\subsection{Ablation Study}
As shown in Table~\ref{tab:ablation_study}, we perform several ablation studies on the AOKVQA~\cite{aokvqa} and MMBench~\cite{mmbench} to validate the effectiveness of each design in \frameworkname.
%The experiments are conducted using LLaVA-1.5-7B~\cite{llava-1.5} under three pruning ratios, \textit{i.e.,} 10\%, 20\% and 30\%.
For each pruning ratio with LLaVA-1.5-7B~\cite{llava-1.5}, we first establish a baseline that utilizes all vision-language (VL) tokens to compute cosine similarity for redundant layer localization and pruning (\textit{i.e.,} straightforwardly extending the approach in Sec.~\ref{sec:layer_pruning_4_llm} to the VL domain). 
From the table, we observe that this approach accelerates LVLMs at the cost of significant performance degradation (\textit{e.g.,} with 3\% and 4.2\% performance loss on AOKVQA and MMBench under the 30\% pruning ratio).

Then, we assess the impact of Token Importance Scores (TIS) by employing them to identify important VL tokens, which are then used to localize less important layers.
The results demonstrate that TIS consistently outperforms the baseline across three pruning ratios.
Moreover, we evaluate the contributions of Subspace-Compensated Pruning (SCP) by applying it on the baseline.
From the table, it can be observed that SCP can significantly boost the performance of the baseline, \textit{e.g.,} achieving a +1.9 performance gain on MMBench under the 30\% pruning rate.
Finally, when we integrate TIS with SCP, \textit{i.e.,} our \noindentframework,
the performance can be further enhanced by a large margin.
Notably, as shown in the last line of the table, our \modelname preserves 99.7\% of the full model's performance while accelerating it by 1.25$\times$.
From the above experiments, we find that each component of \frameworkname plays a crucial role in pruning LVLMs. More importantly, these components operate complementarily, and their combination results in a better trade-off between performance and efficiency.

\begin{table}[t]
\centering
\caption{
The effects of different settings for each component of our \noindentframework. $*$ denotes that the methods need to train the LVLMs.
}
\label{tab:different_model_settings}

\begin{tabular}{lccc}
\toprule
\textbf{}  &   \textbf{AOKVQA} & \textbf{MMBench} & \textbf{Speed}    \\
\midrule
\textbf{\textit{Token Selection:}} \\
Spatial Sampling~\cite{Llava-prumerge} & 73.9 & 58.6 & 6.11 \\
$\gamma$-MOD$*$~\cite{gamma-mod} & 75.1 & 61.2 & 6.11 \\
TIS (Ours) & 76.4 & 62.9 & 6.11 \\
\midrule
\textbf{\textit{Layer Localization:}} \\
ARank~\cite{gamma-mod} & 75.2 & 61.7 & 5.65 \\
ROE-Router$*$~\cite{roe-llava} & 74.8 & 61.8 & 5.65 \\
TIS+Cosine (Ours) & 76.1 & 62.5 & 5.65 \\
\midrule
\textbf{\textit{Feature Gap Mitigation:}} \\
ROE-Adapter$*$~\cite{roe-llava} & 76.3 & 63.9 & 5.10 \\
Fine-tuning$*$ & 76.7 & 63.5 & 5.10 \\
SCP (Ours) & 77.6 & 64.1 & 5.10 \\
\bottomrule
\end{tabular}
\end{table}

\subsection{Different Model Settings}

In this section, we explore several alternative model settings for each component of \noindentframework. In Table~\ref{tab:different_model_settings}, the related experiments are conducted using LLaVA-1.5-7B~\cite{llava-1.5} on the AOKVQA~\cite{aokvqa} and MMBench~\cite{mmbench}.

We first evaluate the effectiveness of our Token Importance Score (TIS) by comparing it with two different \textit{token selection} methods. Specifically, we use a rule-based method, \textit{i.e.,} Spatial Samping~\cite{Llava-prumerge} and a training-based method, \textit{i.e.,} $\gamma$-MOD~\cite{gamma-mod} as baselines. From the table, it can be observed that our TIS achieves better performance than these methods.

Then, we conduct experiments to compare with different \textit{layer localization} methods. In particular, ARank~\cite{gamma-mod} leverages the rank of the attention matrix to evaluate layer redundancy, whereas ROE-Router~\cite{roe-llava} trains a routing mechanism to identify less critical layers. 
Note that the Cosine refers to the cosine similarity.
The results demonstrate that our approach (\textit{i.e.,} TIS+Cosine) achieves superior localization results and outperforms these methods.

Finally, we compare with different \textit{feature gap mitigation} methods, \textit{i.e.,} ROE-Adapter~\cite{roe-llava} and Fine-tuning (the pruned model).
From the table, it can be seen that our Subspace-Compensated Pruning (SCP) also achieves competitive results compared to these training-based methods, further demonstrating the effectiveness of SCP in bridging the feature gaps.

\subsection{Visualizations and Qualitative Analyses}

\subsubsection{Visualization Analyses}

\begin{figure}[t]
  \centering
  \includegraphics[width=1\linewidth]{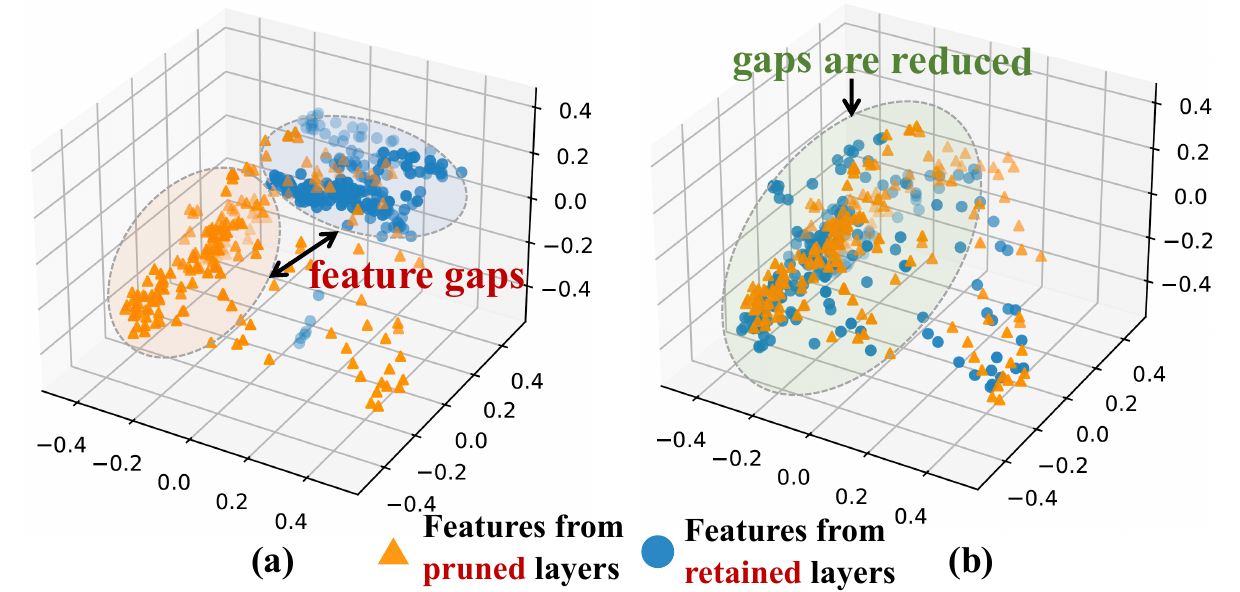}
  \caption{Visualization results of Subspace-Compensated Pruning (SCP). (a) Features from pruned and retained layers \textit{before} using SCP. (b) Features from pruned and retained layers \textit{after} using SCP. The results show that SCP effectively mitigates the feature gaps.}
  \label{fig:scp}
\end{figure}

The core objective of Subspace-Compensated Pruning (SCP) is to restore the performance of pruned LVLMs by bridging the feature gaps between pruned and retained layers. In this section, we conduct experiments to assess whether SCP effectively reduces these feature gaps. In particular, we randomly select 200 data instances from LLaVA-665k~\cite{llava-1.5} and extract their features from both pruned and retained layers. As shown in Fig.~\ref{fig:scp}(a), the features from pruned and retained layers exhibit significant gaps prior to applying SCP. In Fig.~\ref{fig:scp}(b), we observe that these gaps are substantially reduced after applying SCP, demonstrating the effectiveness of SCP in mitigating inter-layer feature gaps.

\subsubsection{Qualitative Results}

\begin{figure}[t]
  \centering
  \includegraphics[width=1\linewidth]{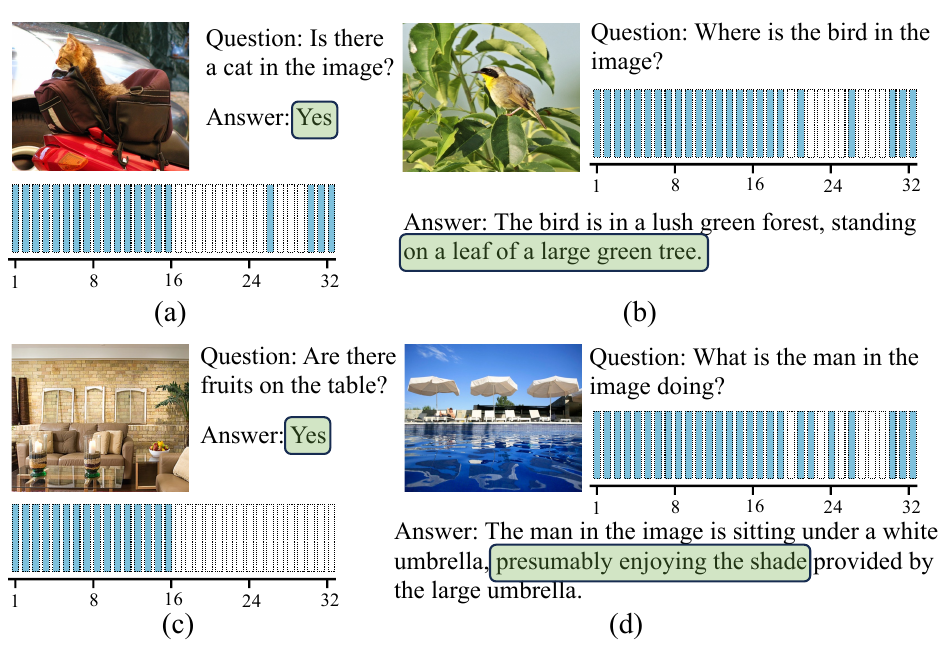}
  \caption{Qualitative results of \noindentframework. The retained layers are depicted in blue.
The results demonstrate that \frameworkname can preserve the capacity of LVLMs even when nearly half of the layers are removed.
  }
  \label{fig:qual_results}
\end{figure}

To further analyze the proposed \noindentframework, we visualize the layer pruning results on LLaVA-1.5-7B~\cite{llava-1.5}. As shown in Fig.~\ref{fig:qual_results}, the multi-modal perception and reasoning capabilities of LVLMs are preserved after pruning layers using \noindentframework. In Fig.~\ref{fig:qual_results}(a) and (c), it can be found that nearly half of the model layers (\textit{i.e.,} 16 layers) can be pruned while the LVLM still correctly answers the given questions. From Fig.~\ref{fig:qual_results}(b) and (d), we observe that many layers of the model can be dropped even when the questions require detailed image understanding and commonsense reasoning.
These results indicate that \frameworkname can effectively prune redundant layers and accelerate the LVLMs without compromising their functionality.

\section{Conclusion}

In this paper, we focus on pruning LVLMs in a training-free manner. We empirically demonstrate that pruning layers in LVLMs is challenging due to the substantial redundancy in
VL tokens and feature gaps between layers. Based on a series of meaningful insights, we propose a new \frameworkname (\noindentmodel) framework that can effectively localize redundant layers using important VL tokens and restore the performance of pruned LVLMs by bridging the layer-wise feature gaps. Meanwhile, \frameworkname is training-free and highly compatible, offering an efficient compression paradigm that can be applied to a wide variety of LVLMs.
We expect that our work can serve as a strong baseline and inspire the VL community to explore more effective and efficient solutions for compressing LVLMs.

\begin{acks}
This work is supported in part by the National Natural Science Foundation of China(62476226).
\end{acks}

%\clearpage
\bibliographystyle{ACM-Reference-Format}
\balance
\bibliography{sample-base}

\end{sloppypar}
\end{document}